\documentclass[11pt]{article}
\usepackage[preprint]{acl}
\usepackage{times}
\usepackage{latexsym}
\usepackage{microtype}
\usepackage{graphicx}
\usepackage{float}
\usepackage{placeins}
\usepackage{booktabs}
\usepackage{multirow}
\usepackage{array}
\usepackage{tabularx}
\usepackage{amsmath}
\usepackage{amssymb}
\usepackage{amsthm}
\usepackage{enumitem}
\usepackage{tikz}
\usetikzlibrary{arrows.meta,positioning,fit,shapes.geometric}
\usepackage{url}

\newtheorem{definition}{Definition}
\newtheorem{property}{Property}
\newtheorem{assumption}{Assumption}

\newcommand{\Actions}{\ensuremath{\mathcal{A}}}
\newcommand{\Feasible}{\ensuremath{\mathcal{F}}}
\newcommand{\Budget}{\ensuremath{\mathcal{B}}}

\newcommand{\LCV}{\ensuremath{\mathrm{LCV}}}
\newcommand{\Repair}{\ensuremath{\Omega}}
\newcolumntype{Y}{>{\raggedright\arraybackslash}X}
\title{Recall Isn’t Enough: Bounding Commitments in Personalized Language Systems}

\author{
\textbf{Rui Tang\textsuperscript{1}} \quad
\textbf{Yichi Zhang\textsuperscript{2}}\thanks{Corresponding author: zhangyichi@stern.nyu.edu} \quad
\textbf{Xi Chen\textsuperscript{2}} \quad
\textbf{Chen Dong\textsuperscript{3}}
\\
\textbf{Youwei Yang\textsuperscript{4,5}} \quad
\textbf{Yumeng Shen\textsuperscript{6}} \quad
\textbf{Qiangqiang Liu\textsuperscript{7}}
\\[0.5ex]
\textsuperscript{1}OpenAsk \quad
\textsuperscript{2}Stern School of Business, New York University \quad
\textsuperscript{3}Bank of Hebei
\\
\textsuperscript{4}Lingnan College, Sun Yat-sen University \quad
\textsuperscript{5}School of Economics, Xiamen University \quad
\textsuperscript{6}BitMart \quad
\textsuperscript{7}Binance
}

\begin{document}
\frenchspacing
\raggedbottom
\maketitle

\begin{abstract}
Long-context and memory systems usually treat personalization as a recall problem. In practice, many failures occur later, when a system commits: it turns noisy hints into hard constraints, drops rare witnesses, forgets downstream obligations, or answers despite infeasibility. We introduce \emph{Contract-Bounded Evidence Activation} (CBEA) with \emph{Lexicographic Commitment Validation} (LCV). CBEA activates a bounded evidence set using typed coverage, tail witnesses, and consequence debt; LCV validates structured commitments before prose and routes infeasible states to repair, abstention, or recontract. Across 360 fixtures and three generation backends, CBEA+LCV reaches zero failures within validator scope at 0.49--0.60 availability over attempted runs. Raw and long-context baselines with the same LCV gate reach zero only at 0.003--0.092. A shadow oracle diagnostic marks the limit: CBEA+LCV recalls 0.012 of uncompiled visible facts, while raw recalls 0.53. The result is a bounded operating point: explicit commitment control and 74--75\% lower median input payload, not universal memory dominance.
\end{abstract}

\section{Introduction}

Long-horizon personalization must preserve confirmed user constraints while adapting to noisy, evolving state. Fluent responses can still drop hard constraints, overweight recent context, lose rare witnesses, or continue when no response is compatible with the contract. We call these \emph{runtime control failures}: errors that occur after evidence is collected but before a commitment is safely realized. The standard alternative---stuffing the raw user history into the prompt---asks the model to rediscover which facts matter at every turn. The same sentence may be reinterpreted across turns, hard constraints flatten into soft preferences, and rare evidence disappears unless the runtime keeps an explicit obligation model.

This failure mode is distinct from ordinary memory failure. A system may retrieve the relevant sentence and still make the wrong commitment if it treats a soft hint as a hard constraint, ignores a required witness, or fails to recognize that the feasible set is empty. Conversely, a system may be fluent and useful while remaining outside any claim we can verify. The central question is therefore not only what context to show a model, but what commitments the runtime is allowed to realize after that context is selected.

We address this with runtime control bounded by contracts: runtime decisions are constrained by explicitly confirmed structured commitments covered by validators. Clarification targets missing information with high downstream impact; compilation converts confirmed information into hard predicates, evidence units, mutable state, and required coverage; CBEA selects bounded evidence; LCV filters structured commitments before prose; and infeasible states route to repair, abstention, or recontract rather than to new commitments.

The paper deliberately separates two boundaries. Inside the compiled contract, the runtime can validate structured commitments against explicit predicates and coverage requirements. Outside that boundary, uncompiled facts remain ordinary context: the method can miss them, and we measure that exclusion rather than hiding it. This is why the evaluation reports both validator-covered control failures and a shadow-oracle diagnostic over visible facts that were not compiled into the contract.

The paper makes four contributions:
\begin{enumerate}[leftmargin=1.5em,itemsep=1pt,parsep=0pt,topsep=2pt]
    \item CBEA, a budgeted evidence activation objective over local relevance, contract coverage, tail witness retention, consequence debt, and overpersonalization penalties;
    \item LCV, a lexicographic validation rule whose invariant is attached to structured commitments covered by validators rather than unrestricted prose;
    \item a matched evaluation across nine variants and 360 fixtures, covering hard constraints, coverage, witnesses, infeasible continuation, consequences, and repair, plus a long history payload diagnostic built from prompts selected by CBEA;
    \item an audit of 90 cases by model judges over six fidelity dimensions, combined with privacy protected production aggregates as diagnostic evidence rather than causal outcome evidence.
\end{enumerate}

\section{Related Work and Positioning}

\paragraph{Long horizon.}
Recent benchmarks already study long-horizon personalization and contextual preference inference. HorizonBench tracks evolving preferences over long simulated histories \cite{horizonbench2026}; CUPID evaluates contextualized alignment from interaction histories \cite{cupid2025}. Our target is not preference inference alone, but runtime control after evidence exists.

\paragraph{Memory.}
Memory systems address recall under context pressure. TiMem proposes temporal hierarchical memory consolidation \cite{timem2026}, while OP-Bench shows that memory can be misused through irrelevance, repetition, and sycophancy \cite{opbench2026}. We also include a selector-level diagnostic against a classic MMR selector \cite{carbonell1998mmr}; unlike CBEA, MMR has no typed coverage, tail witness, or consequence debt terms. We therefore treat evidence activation as a constrained runtime decision before validation and realization.

\paragraph{Clarification.}
Prior work studies clarification for preference elicitation \cite{google_clarify_2025,selm2024}, profile compression \cite{personax2025,difference_aware_2025}, and long context limits \cite{context_length_2025}. These motivate moving runtime truth from raw profile text to explicit artifacts.

\paragraph{Prompt compression and RAG.}
Compression and retrieval baselines reduce context cost, but they do not by themselves define which constraints are hard, which evidence is mandatory, or what the system must do when no feasible answer exists. Our comparison therefore includes raw history, summary, dense retrieval, long-context prompting, and tool/memory baselines, but the main claim concerns the control interface placed after evidence selection: a structured commitment is either covered by validators or outside the guarantee.

\paragraph{Verification.}
Contract-grounded planning separates retrieval, typed constraints, verification, repair, and abstention \cite{contract2plan2026}. Our setting differs: constraints are confirmed by the user, state evolves, evidence is incomplete, and validated commitments are realized through language. We borrow the boundary that safety claims apply only to structured commitments covered by validators.

\paragraph{Position.}
Our closest comparison is not personalization memory alone, but systems that separate retrieval, constraint representation, validation, and repair. Unlike previous work, our contribution is their integration into a runtime control method for noisy long-horizon personalized language systems.

\section{Formal Runtime Model}
\label{sec:formal-model}

The paper's central object is a runtime over structured commitments. Let $x_{0:t}$ be a sequence of noisy user observations, including forms, labels, free text, and previous turns. The runtime maintains a confirmed hard contract $h_t$, mutable state $u_t$, a shared evidence pool $E_t$, a required-coverage set $R_t$, and a structured action space $\Actions$.

\begin{definition}[Confirmed hard contract]
A hard contract is a set
\[
h_t=\{(\phi_j,\pi_j)\}_{j=1}^{m_t},
\]
where each $\phi_j:\Actions \rightarrow \{0,1\}$ is a machine-checkable hard predicate and $\pi_j$ is a provenance record. Only explicitly confirmed predicates belong to $h_t$. Inferred or unresolved constraints remain outside the immutable contract.
\end{definition}

\begin{definition}[Evidence pool]
The evidence pool is a finite set
\[
E_t=\{e_i=(z_i,d_i,\rho_i,\tau_i,\kappa_i)\}_{i=1}^{n_t},
\]
where $z_i$ is an evidence unit, $d_i$ its dimension, $\rho_i$ provenance metadata, $\tau_i$ a tail witness indicator, and $\kappa_i$ an activation cost. Selected evidence units, not raw profile text, are the operational interface to generation.
\end{definition}

\begin{definition}[Required coverage set]
The runtime derives a set
\[
R_t=\Gamma(h_t,u_t,c_t)
\]
of required evidence dimensions for the current turn or result. Requirements
may come from confirmed hard predicates, required evidence fields, consequence
debt, local scene obligations, or no-feasible checks. Let
$M_t\in\{0,1\}^{|E_t|\times |R_t|}$ be an evidence requirement coverage matrix,
where $M_{ir}=1$ means evidence unit $e_i$ covers requirement $r$.
\end{definition}

\paragraph{CBEA.}
CBEA selects evidence through a budgeted objective rather than top-$k$ retrieval.
For a candidate evidence subset $Z\subseteq E_t$, define
\[
\begin{aligned}
J_t(Z)={}&
\lambda_r\,\mathrm{Rel}(Z,c_t)
+\lambda_c\,\mathrm{Cov}(Z,R_t)\\
&+\lambda_w\,\mathrm{Tail}(Z)
+\lambda_d\,\mathrm{Debt}(Z,u_t)\\
&-\lambda_o\,\mathrm{Over}(Z,c_t),
\end{aligned}
\]
where $\mathrm{Rel}$ measures local relevance, $\mathrm{Cov}$ measures
requirement coverage, $\mathrm{Tail}$ rewards retention of rare but decisive
tail witnesses, $\mathrm{Debt}$ rewards evidence needed for downstream
obligations, and $\mathrm{Over}$ penalizes irrelevant or intrusive
personalization. Requirement coverage is computed from $M_t$ using a short
coverage indicator:
\[
\begin{aligned}
\mathrm{Cov}(Z,R_t)&=\textstyle\sum_{r\in R_t}w_r\,\eta_r(Z),\\
\eta_r(Z)&=\mathbf{1}[\exists e_i\in Z: M_{ir}=1].
\end{aligned}
\]
CBEA chooses
\[
Z_t^\star \in \operatorname*{arg\,max}_{Z\subseteq E_t} J_t(Z)
\quad \text{s.t.}\quad
\sum_{e_i\in Z}\kappa_i\leq \Budget_t .
\]
In implementation, this objective can be solved approximately with greedy
budgeted coverage. A reserved budget
$\Budget_t=\Budget_t^{main}+\Budget_t^{tail}$ lets the runtime protect required
tail witnesses separately from ordinary evidence with high relevance. The
corresponding ablations remove coverage, tail reservation, or consequence debt
terms.

\paragraph{LCV.}
A candidate generator constructs a finite set $A_t \subseteq \Actions$ of
structured commitments from $Z_t^\star$. For each candidate, LCV computes a
violation vector
\[
\begin{aligned}
\nu_t(a)=\big(&\nu_h(a,h_t),\nu_c(a,Z_t^\star,R_t),\\
&\nu_0(a),-S(a;u_t,Z_t^\star,c_t)\big),
\end{aligned}
\]
where $\nu_h$ counts hard predicate failures covered by validators,
$\nu_c$ counts missing required evidence coverage, $\nu_0$ indicates
commitment emission when the runtime diagnoses no feasible
candidate covered by validators, and $S$ is a soft utility score over mutable state
and local context. LCV ranks candidates by
lexicographic minimization:
\[
\tilde{a}_t \in
\operatorname*{arg\,lexmin}_{a\in A_t}\nu_t(a).
\]
For readability, write
$\bar{\nu}_{h,t}(a)=\nu_h(a,h_t)$ and
$\bar{\nu}_{c,t}(a)=\nu_c(a,Z_t^\star,R_t)$. The feasible set is
\[
\begin{aligned}
\Feasible_t=\{a\in A_t:\;&
\bar{\nu}_{h,t}(a)=0,\\
&\bar{\nu}_{c,t}(a)=0\}.
\end{aligned}
\]
If $\Feasible_t\neq\varnothing$, the runtime emits
\[
a_t^\star \in
\operatorname*{arg\,max}_{a\in\Feasible_t}
S(a;u_t,Z_t^\star,c_t),
\]
realizes it and updates mutable state:
\[
y_t=G_\psi(a_t^\star,u_t,Z_t^\star,c_t),\;
u_{t+1}=U(u_t,a_t^\star,x_{t+1}).
\]

If $\Feasible_t=\varnothing$, the runtime must not emit a commitment as if it
were feasible. LCV diagnoses an infeasibility reason
such as missing evidence, contract conflict, unsupported commitment, or
validator failure. Let $\delta_t$ denote this reason. The runtime calls a
restricted repair operator
\[
o_t=\Repair(\delta_t,h_t,u_t,Z_t^\star,c_t),
\]
where $o_t$ is repair, abstention, fallback, or explicit recontract. In our
implementation, missing evidence maps to clarification, contract conflict to
recontract, and unsupported or validator failing candidates to abstention or
fallback. These acts do not carry commitments and are scored separately from
valid structured commitments. Appendix~\ref{app:lcv-repair-case} gives an
illustrative LCV repair routing sketch.

\begin{assumption}[Structured emission covered by validators]
The validator exactly covers the confirmed predicates in $h_t$ for parseable structured commitments. Repair, abstention, and recontract acts do not carry commitments: they may ask for missing information, decline to continue, or request explicit contract correction, but they must not smuggle a new commitment as advice. Timeout, no output, parse failure, partial output, and blank output are not counted as safe emissions; they are system level failures or unevaluable states with explicit denominators.
\end{assumption}

\begin{property}[Emission boundary covered by validators]
Under exact validator coverage for predicates in $h_t$, any emitted structured commitment $a_t^\star$ satisfies all confirmed hard predicates in $h_t$. When $\Feasible_t=\varnothing$, no structured commitment is emitted; the runtime is restricted to repair, abstention, or explicit recontract. This invariant is distinct from the oracle-level violation metrics used in the benchmark.
\end{property}

This property is a claim boundary, not a theorem over all natural language implications. The guarantee is attached to $a_t^\star$, and the surface realizer $G_\psi$ remains a separately evaluated layer because prose can introduce unsupported implications even when the underlying commitment is valid.

\begin{table}[t]
\centering
\small
\begingroup
\renewcommand{\arraystretch}{1.14}
\setlength{\tabcolsep}{4pt}
\begin{tabularx}{\linewidth}{@{}lYY@{}}
\toprule
\textbf{Object} & \textbf{Runtime role} & \textbf{Failure measured} \\
\midrule
$h_t$ & Confirmed hard contract & Hard constraint drift or false hardening \\
\addlinespace[0.2em]
$E_t$ & Shared evidence pool & Witness loss and coverage failure \\
\addlinespace[0.2em]
$u_t$ & Mutable turn state & Trajectory and consequence discontinuity \\
\addlinespace[0.2em]
$R_t$ & Required coverage set & Missing required dimensions \\
\addlinespace[0.2em]
$M_t$ & Evidence requirement matrix & Coverage accounting error \\
\addlinespace[0.2em]
$Z_t$ & Activated evidence & Prompt budget omission and tail loss \\
\addlinespace[0.2em]
$a_t$ & Structured commitment & Invalid feasible set membership \\
\addlinespace[0.2em]
\LCV & Lexicographic validator & Contract violation covered by validators \\
\addlinespace[0.2em]
\Repair & Repair/recontract operator & Invalid infeasible state handling \\
\addlinespace[0.2em]
$G_\psi$ & Surface realizer & Prose realization error \\
\bottomrule
\end{tabularx}
\endgroup
\caption{The paper is organized around runtime objects. Evaluation metrics are derived from these objects rather than from generic fluency or engagement.}
\label{tab:objects}
\end{table}

\section{CBEA and LCV Runtime Algorithm}

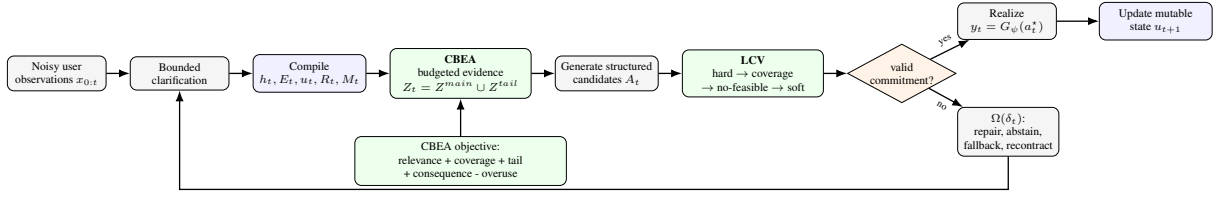
\begin{figure*}[t]
\centering
\resizebox{\textwidth}{!}{%
\begin{tikzpicture}[
  font=\scriptsize,
  node distance=5mm and 5mm,
  box/.style={draw, rounded corners, align=center, minimum width=21mm, minimum height=7mm, fill=gray!8},
  state/.style={draw, rounded corners, align=center, minimum width=24mm, minimum height=8mm, fill=blue!6},
  method/.style={draw, rounded corners, align=center, minimum width=30mm, minimum height=10mm, fill=green!7},
  gate/.style={draw, diamond, aspect=1.85, align=center, fill=orange!10, inner sep=0.6pt},
  arrow/.style={-Latex, thick}
]
\node[box] (input) {Noisy user\\observations $x_{0:t}$};
\node[box, right=of input] (clarify) {Bounded\\clarification};
\node[state, right=of clarify] (compile) {Compile\\$h_t,E_t,u_t,R_t,M_t$};
\node[method, right=of compile] (cbea) {\textbf{CBEA}\\budgeted evidence\\$Z_t=Z^{main}\cup Z^{tail}$};
\node[box, right=of cbea] (candidate) {Generate structured\\candidates $A_t$};
\node[method, right=of candidate] (lcv) {\textbf{LCV}\\hard $\rightarrow$ coverage\\$\rightarrow$ no-feasible $\rightarrow$ soft};
\node[gate, right=of lcv] (validate) {valid\\commitment?};
\node[box, above right=4mm and 6mm of validate] (realize) {Realize\\$y_t=G_\psi(a_t^\star)$};
\node[box, below right=4mm and 6mm of validate] (repair) {$\Repair(\delta_t)$:\\repair, abstain,\\fallback, recontract};
\node[state, right=9mm of realize] (update) {Update mutable\\state $u_{t+1}$};
\node[method, below=8mm of cbea, minimum width=45mm] (obj) {CBEA objective:\\relevance + coverage + tail\\+ consequence - overuse};
\draw[arrow] (input) -- (clarify);
\draw[arrow] (clarify) -- (compile);
\draw[arrow] (compile) -- (cbea);
\draw[arrow] (cbea) -- (candidate);
\draw[arrow] (candidate) -- (lcv);
\draw[arrow] (lcv) -- (validate);
\draw[arrow] (validate) -- node[above, sloped, font=\tiny] {yes} (realize);
\draw[arrow] (validate) -- node[below, sloped, font=\tiny] {no} (repair);
\draw[arrow] (realize) -- (update);
\draw[arrow] (repair.south) -- ++(0,-7mm) -| (clarify.south);
\draw[arrow] (obj.north) -- (cbea.south);
\end{tikzpicture}
}
\caption{Runtime control with validator gates, CBEA, and LCV. Evidence is activated under budget, commitments are validated before prose realization, and infeasible states route to repair, abstention, fallback, or recontract.}
\label{fig:runtime}
\end{figure*}

Figure~\ref{fig:runtime} shows the method. The model is not asked to rediscover the user contract from raw background text on every turn; the runtime compiles constraints, activates evidence, validates commitments, and only then realizes language.

\paragraph{Compilation.}
Clarification is bounded information acquisition, not open interviewing. Compilation converts confirmed answers into $h_t$, extracts evidence units into $E_t$, initializes $u_t$, derives $R_t$, and populates $M_t$, keeping inferred soft preferences outside the hard contract.

\paragraph{Activation.}
Evidence activation is constrained selection, not unconditional memory retrieval. CBEA adds coverage of $R_t$, tail witness reservation, consequence debt evidence, and overpersonalization penalties to local relevance. Turn activation prioritizes latency; result activation prioritizes synthesis and consequence preservation.

\paragraph{Validation.}
The runtime validates structured commitments before language. Hard predicates are lexicographically prior to evidence coverage, and both precede soft utility. More evidence cannot compensate for violating an explicitly confirmed boundary. If every candidate violates $h_t$ or lacks required coverage, the runtime forbids commitments that claim feasibility and routes to $\Repair(\delta_t)$.

\paragraph{Realization.}
A valid commitment can still be phrased ambiguously or with unsupported implications, so surface realization remains a language generation problem evaluated separately (\S\ref{sec:model-judge-audit}).

\paragraph{Implementation.}
CBEA is not a learned preference model. In our experiments, $R_t$ and $M_t$ are constructed from fixture schemas and runtime contract rules: confirmed hard predicates induce mandatory requirements, fixture witnesses mark covered requirements, and consequence debt obligations follow from confirmed state transitions. Oracle labels $(h^\star,E^\star,\Feasible^\star)$ are used only for scoring. They are not inputs to clarification, contract compilation, CBEA selection, candidate generation, LCV validation, repair routing, or prose realization. CBEA weights are fixed design priors $(\lambda_r,\lambda_c,\lambda_w,\lambda_d,\lambda_o)=(1,2,2,1,1)$, not learned, tuned on a development set, or tuned after the fact for metric maximization; they encode the priority that coverage and tail witnesses dominate ordinary relevance. Candidates share a schema (commitment type, evidence ids, claimed predicates, required slots, repair status, surface text); validators cover hard contract consistency, required evidence coverage, infeasible emission, and repair reason validity. These choices make the method auditable but also bound the claim: results test this fixed instantiation, not an optimized learned selector. Appendix~\ref{app:method-details} lists the reporting schema.

\section{Evaluation and Benchmark Protocol}

The evaluation follows the formal objects in Table~\ref{tab:objects}. We use narrative decision support as a hostile stress test because it combines noisy self report, delayed consequences, household or social spillovers, long free text background, and recovery after drift. The benchmark does not claim to measure real world decision quality.

\subsection{Fixture Construction}

Fixtures derived from real users are synthetic or composite cases derived from
aggregate failure patterns, never raw histories. The key transformation is from
observed failure families to controlled fixtures with known evaluator labels:
confirmed predicates $h^\star$, evidence witnesses $E^\star$, expected feasible set $\Feasible^\star$, and allowed repair or abstention behavior. The construction pipeline---aggregate failure patterns~$\rightarrow$ abstract stress mechanism~$\rightarrow$ synthetic/composite profile rewrite~$\rightarrow$ structured labels~$\rightarrow$ controlled offline fixture---is detailed in Appendix~\ref{app:fixture-boundaries}, which also reports the exported fixture buckets and stress surface mapping.

The fixtures cover seven stress surfaces: false hardening, hidden exception, witness drop, infeasible continuation, consequence debt, overpersonalization, and surface mismatch. Six are exported as primary fixture buckets in the matched manifest; overpersonalization is retained as an activation penalty rather than a separate bucket (Appendix~\ref{app:fixture-boundaries}).

\subsection{Metrics}

Let $\mathcal{D}$ be a controlled benchmark set, $\mathcal{D}_{emit}$ the subset with parseable emitted structured commitments, and $\mathcal{D}_{0}$ the subset whose oracle feasible set is empty. Let $q_f=1$ when fixture $f$ emits a parseable structured commitment and $q_f=0$ otherwise. Structured output availability is
\[
\mathrm{Avail} =
\frac{1}{|\mathcal{D}|}\sum_{f\in\mathcal{D}} q_f .
\]
Let $v_f=1$ when $\hat{a}_f$ violates at least one oracle predicate in $h_f^\star$, and $v_f=0$ otherwise. Oracle hard constraint violation rate is reported over parseable emitted commitments:
\[
\mathrm{OHCVR}=
\frac{1}{|\mathcal{D}_{emit}|}
\sum_{f\in\mathcal{D}_{emit}} v_f .
\]
This oracle level metric is stricter than the invariant covered by validators above: it can count hidden, uncompiled, or incorrectly realized fixture predicates that were not certified by $V_h$.
Let $n_f=1$ when the emitted control act carries a commitment in an infeasible fixture, and $n_f=0$ for repair, abstention, or recontract. The infeasible emission rate is
\[
\mathrm{NFER}=
\frac{1}{|\mathcal{D}_{0}|}
\sum_{f\in\mathcal{D}_{0}} n_f ,
\]
where a commitment output in this setting is an infeasible emission failure. Evidence coverage failure and consequence continuity failure are computed analogously from fixture labels. Repair correctness is computed over $\mathcal{D}_{repair}$, the subset of fixtures whose evaluator label expects repair, abstention, or recontract behavior (including but not limited to $\mathcal{D}_0$); it scores 1 when the emitted control act matches the expected repair category without making a commitment. Surface realization is evaluated separately in the blinded model judge audit, where judges score surface coherence from rubric labels rather than an automatic proxy. The surface mismatch fixture bucket still contributes its structured contract, coverage, witness, consequence, infeasible emission, and repair labels to the automatic metrics in Table~\ref{tab:offline-results}; only surface coherence itself is not scored by an automatic proxy.

Metrics requiring a structured commitment are reported on $\mathcal{D}_{emit}$ with denominator shown. Aggregate system pass rates include timeout, no output, parse failure, partial output, blank output, and unrepaired error states as failures rather than silently removing them (Appendix~\ref{app:denominators}).

\subsection{Baselines and Ablations}

The run includes two isolating variants: validator only separates validation
and repair from evidence activation, and runtime without CBEA keeps structured
state and repair while replacing CBEA with evidence selection without tail reserve. We use
these rows as targeted component diagnostics for the activation and validation
interfaces. Appendix~\ref{app:method-details}, Table~\ref{tab:baselines}
summarizes the nine comparison interfaces.

\paragraph{Model reporting.}
Controlled comparisons report model identifiers, decoding settings, context/generation budgets, retrieval budgets, tool availability, and memory availability. Production traffic is not used for model comparison because deployed provider/model mix changes over time and is only operational context.

\paragraph{Reproducibility.}
We provide an anonymized artifact that preserves privacy and includes the
fixture generator, 360-fixture benchmark, runtime harness with CBEA selector and
LCV validators, released model output result tables, metric summarizers, paired
and case cluster bootstrap scripts, selector level MMR diagnostic,
backend sensitivity configs, and privacy boundary checks. It supports
reproduction of Table~\ref{tab:offline-results}, the long history payload
diagnostic, targeted ablations, backend sensitivity operating points, and
model judge bootstrap summaries; it contains no raw user histories or production
text, and production evidence is released only as aggregate counts.

\section{Controlled Offline Experiments}
\label{sec:offline-experiments}

We ran 360 synthetic and composite fixtures across the nine variants (Appendix~\ref{app:method-details}, Table~\ref{tab:baselines}). All non-oracle variants used MiniMax-M2.7 in a matched offline run with temperature 0.2, a 2200-token cap, up to three JSON parse retries, and a 180-second timeout. The final run contains 3,240 attempted and evaluable rows. Oracle evidence is deterministic and included only as a reference point.

\begin{table}[H]
\centering
\scriptsize
\begingroup
\renewcommand{\arraystretch}{1.05}
\setlength{\tabcolsep}{3pt}
\begin{tabular}{lrrrrr}
\toprule
\textbf{Method} & \textbf{Att.} & \textbf{Inv.} & \textbf{Struct.} & \textbf{Repair} & \textbf{Cost} \\
\midrule
Raw prompt stuffing  & 360 & 0 & \textbf{0.7833} & 0.4571 & 3218 \\
Summarized profile   & 360 & 0 & 0.4500 & 0.6786 & 3395 \\
Dense retrieval RAG  & 360 & 0 & 0.7333 & 0.3571 & 2612 \\
Long-context LLM     & 360 & 0 & 0.7278 & 0.4786 & 3279 \\
Tool/memory agent    & 360 & 0 & 0.7500 & 0.4643 & 3189 \\
Validator-only       & 360 & 0 & 0.4028 & 1.0000 & \textbf{2778} \\
Runtime w/o CBEA     & 360 & 0 & 0.5000 & 1.0000 & 3542 \\
CBEA + LCV runtime   & 360 & 0 & 0.5000 & 1.0000 & 3178 \\
\midrule
Oracle (ref.)        & 360 & 0 & 0.6111 & 1.0000 & -- \\
\bottomrule
\end{tabular}
\endgroup
\caption{Matched offline operating points over 360 synthetic/composite fixtures (MiniMax-M2.7). Att.: attempted runs; Inv.: invalid; Struct.: structured commitment availability; Repair: $\mathcal{D}_{repair}$ correctness; Cost: average prompt cost units. Appendix~\ref{app:covered-detail} reports OHCVR/ECF/Wit./Cons./NFER for all nine methods; Appendix~\ref{app:shadow-oracle} reports the uncompiled context boundary.}
\label{tab:offline-results}
\end{table}

\begin{figure}[H]
\centering
\resizebox{0.82\columnwidth}{!}{%
\begin{tikzpicture}[
  font=\footnotesize,
  x=9cm, y=4.0cm,
  ungated/.style={circle, fill=gray!55, draw=white, line width=0.5pt, inner sep=1.6pt},
  lcvgated/.style={circle, fill=orange!75!black, draw=white, line width=0.5pt, inner sep=1.6pt},
  raw/.style={circle, fill=black, draw=white, line width=0.7pt, inner sep=2.2pt},
  cbea/.style={circle, fill=blue!65!black, draw=white, line width=0.7pt, inner sep=2.2pt},
  oracle/.style={circle, draw=gray!70, fill=white, line width=0.7pt, inner sep=1.8pt},
  axislbl/.style={font=\scriptsize},
  ticklbl/.style={font=\tiny}
]
\foreach \x in {0.4, 0.5, 0.6, 0.7, 0.8} {
  \draw[gray!18, line width=0.25pt] (\x, 0) -- (\x, 0.68);
}
\foreach \y in {0.1, 0.2, 0.3, 0.4, 0.5, 0.6} {
  \draw[gray!18, line width=0.25pt] (0.35, \y) -- (0.88, \y);
}

\draw[gray!70, line width=0.6pt] (0.35, 0) -- (0.88, 0);
\draw[gray!70, line width=0.6pt] (0.35, 0) -- (0.35, 0.68);

\foreach \x in {0.4, 0.5, 0.6, 0.7, 0.8} {
  \draw[gray!70, line width=0.4pt] (\x, 0) -- (\x, -0.010);
  \node[ticklbl, below=1pt] at (\x, 0) {\x};
}
\foreach \y in {0.0, 0.2, 0.4, 0.6} {
  \draw[gray!70, line width=0.4pt] (0.35, \y) -- (0.343, \y);
  \node[ticklbl, left=1pt] at (0.343, \y) {\y};
}

\node[axislbl, below=10pt] at (0.615, 0) {Structured commitment availability};
\node[axislbl, rotate=90, anchor=south] at (0.295, 0.34) {Mean control failure rate};

\draw[->, gray!45, line width=0.6pt] (0.84, 0.55) .. controls (0.86, 0.40) and (0.86, 0.20) .. (0.84, 0.04);
\node[font=\tiny, gray!55!black, anchor=west] at (0.845, 0.30) {better};

\node[ungated] at (0.4500, 0.5746) {};
\node[ungated] at (0.7333, 0.5600) {};
\node[ungated] at (0.7278, 0.2670) {};
\node[ungated] at (0.7500, 0.3889) {};

\node[lcvgated] at (0.4028, 0.5476) {};
\node[lcvgated] at (0.5000, 0.4067) {};

\node[oracle] (orc) at (0.6111, 0.0000) {};
\node[font=\tiny, gray!50!black, above right=-1pt and 0pt] at (orc) {Oracle};

\node[raw] (rawp) at (0.7833, 0.2173) {};
\node[font=\scriptsize\bfseries, above=3pt] at (rawp) {Raw};

\node[cbea] (cbeap) at (0.5000, 0.0000) {};
\node[font=\scriptsize\bfseries, text=blue!65!black, above=3pt] at (cbeap) {CBEA+LCV};

\draw[->, dashed, gray!50, line width=0.5pt] (rawp) to[bend right=15] (cbeap);

\end{tikzpicture}%
}
\caption{Operating point view of Table~\ref{tab:offline-results}. Raw prompt stuffing emits more commitments but has higher control failures on covered surfaces; CBEA+LCV emits fewer commitments but reaches the zero failure operating point within the validator scope. Other baselines (\textcolor{orange!75!black}{orange}: with LCV gates; gray: ungated) are shown as context, not as a universal dominance claim. Dashed arrow: design shift raw$\rightarrow$CBEA+LCV.}
\label{fig:operating-point}
\end{figure}
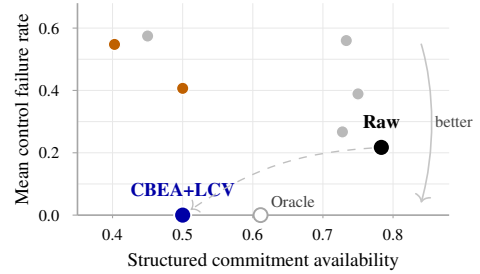

Table~\ref{tab:offline-results} and Figure~\ref{fig:operating-point} show a shift in operating point, not broad dominance. CBEA+LCV emits structured commitments on 0.5000 of fixtures with zero measured failures on covered surfaces; raw prompt stuffing emits on 0.7833 but has failures across every covered surface. The remaining CBEA+LCV fixtures route to repair, abstention, or recontract without making new commitments, with $\mathcal{D}_{repair}$ correctness $=1.0000$. Raw remains competitive on availability because the fixtures expose relevant evidence directly. The runtime contribution is to make constraint, coverage, witness, and consequence behavior \emph{measurable and enforceable} on emitted commitments.

This operating point is intentionally conservative. A system that emits on more cases can look better under a simple availability metric while silently violating hard predicates or losing required evidence. A system that refuses too often can also drive failure rates to zero without solving the activation problem. We therefore read Table~\ref{tab:offline-results} together with Figure~\ref{fig:operating-point}: the target is not the upper-right corner, but a point where emitted commitments remain covered and unrepaired cases are explicitly routed.

Property~1 explains why LCV can suppress infeasible commitments or commitments that violate validators; it does not ensure that the generator sees the right witnesses or consequence obligations. Those surfaces depend on activation: validator-only and runtime without CBEA still lose coverage, witnesses, and consequences (Appendix~\ref{app:covered-detail}). Thus validation alone can refuse infeasible cases, but CBEA supplies the evidence state that makes validation useful.

Targeted ablations (Appendix~\ref{app:targeted-ablations}) isolate validation, repair, and coverage/tail activation; a selector diagnostic (Appendix~\ref{app:mmr-selector}) shows CBEA retains 0.9970 of typed control evidence versus 0.6960 for MMR. Backend sensitivity and the Hy3-preview output budget diagnostic are summarized below; horizon complexity and long history payload diagnostics reuse the same harness.

\begin{figure*}[!t]
\centering
\begin{tikzpicture}[
  font=\footnotesize,
  panel/.style={draw=gray!60, line width=0.4pt},
  bar_raw/.style={fill=gray!45, draw=gray!70, line width=0.2pt},
  bar_long/.style={fill=gray!70, draw=gray!85, line width=0.2pt},
  bar_cbea/.style={fill=blue!60!black, draw=none}
]
\foreach \i/\xshift/\title/\rv/\lv/\cv in {%
  0/0cm/MiniMax-M2.7/0.0361/0.0500/0.4944,%
  1/4.7cm/DeepSeek-V4-Flash/0.0917/0.0778/0.6000,%
  2/9.4cm/GPT-OSS-120B/0.0028/0.0056/0.5056%
} {
  \begin{scope}[xshift=\xshift]
    \draw[panel] (0,0) rectangle (3.9, 3.0);
    \foreach \v in {0.2, 0.4, 0.6} {
      \draw[gray!25, line width=0.25pt] (0, \v/0.65*3) -- (3.9, \v/0.65*3);
    }
    \fill[bar_raw]  (0.55, 0) rectangle (1.35, \rv/0.65*3);
    \fill[bar_long] (1.55, 0) rectangle (2.35, \lv/0.65*3);
    \fill[bar_cbea] (2.55, 0) rectangle (3.35, \cv/0.65*3);
    \node[font=\scriptsize] at (0.95, \rv/0.65*3 + 0.22) {\pgfmathprintnumber[fixed,precision=3]{\rv}};
    \node[font=\scriptsize] at (1.95, \lv/0.65*3 + 0.22) {\pgfmathprintnumber[fixed,precision=3]{\lv}};
    \node[font=\scriptsize\bfseries] at (2.95, \cv/0.65*3 + 0.22) {\pgfmathprintnumber[fixed,precision=3]{\cv}};
    \node[font=\scriptsize, anchor=north] at (0.95, -0.05) {Raw+L};
    \node[font=\scriptsize, anchor=north] at (1.95, -0.05) {Long+L};
    \node[font=\scriptsize, anchor=north] at (2.95, -0.05) {CBEA+L};
    \node[font=\footnotesize\bfseries] at (1.95, 3.3) {\title};
  \end{scope}
}
\foreach \v/\lbl in {0/0.0, 0.2/0.2, 0.4/0.4, 0.6/0.6} {
  \node[font=\scriptsize, anchor=east] at (-0.05, \v/0.65*3) {\lbl};
}
\node[font=\scriptsize, rotate=90] at (-0.85, 1.5) {Attempted run availability};
\end{tikzpicture}
\caption{Backend sensitivity operating points over the matched 360-fixture comparison with LCV gates. Bars show attempted run structured commitment availability for Raw+LCV, Long-context+LCV, and CBEA+LCV; all cells have zero measured failures covered by validators on emitted commitments. Full system cost diagnostics are in Appendix~\ref{app:backend-robustness}.}
\label{fig:backend-availability}
\end{figure*}
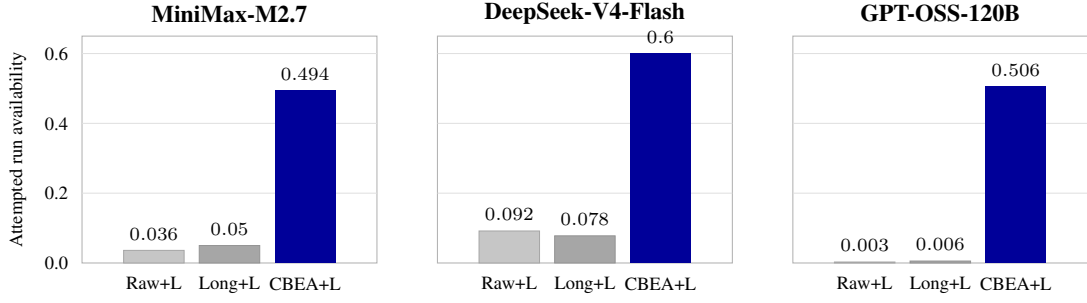

\paragraph{Backend sensitivity.} Across the three backends, CBEA+LCV attempted run availability is 0.4944, 0.6000, and 0.5056; raw and long-context baselines with LCV gates reach the same zero failure level only at 0.0028--0.0917. The pattern preserves direction across two model families and three labs, but remains a sensitivity diagnostic rather than a robustness claim. Hy3-preview is reported separately: at 2200 tokens it does not emit parseable structured commitments, while an 8800-token diagnostic is small and not matched to the 360-fixture protocol.

\paragraph{Activation boundary.} Property~1 suppresses infeasible commitments or commitments that violate validators, but it cannot make missing witnesses or consequence obligations visible. CBEA's role is therefore not a tighter validator; it is a selector whose typed coverage, tail witness, and consequence debt terms put the right evidence units into $Z_t^\star$ before validation runs.

\paragraph{Shadow exclusion boundary.} A shadow oracle diagnostic (Appendix~\ref{app:shadow-oracle}) maps visible but uncompiled facts outside the compiled guarantee. CBEA+LCV is nearly inert on them; raw prompt stuffing retains more. This is not a second success criterion for CBEA+LCV, but the exclusion side of the guarantee. The diagnostic prevents the main zero-failure result from being read as universal memory coverage: CBEA+LCV controls what it compiles, and raw history exposes more uncompiled facts while failing more often on covered commitments. The contribution is a precisely characterized operating point within the validator scope; combining that point with broader uncompiled-context awareness is future work.

\paragraph{Production-scale diagnostics.} A privacy-protected deployment export
(524 attempted sessions over a 720-hour window, 447 in the semantic pool after
operational failures are separated, 6{,}018 evaluable turns) supports deployment
scale and reliability diagnostics only; it is not correctness evidence. Full
denominators are in Appendix~\ref{app:production}.

\section{Blinded Model Judge Fidelity Audit}
\label{sec:model-judge-audit}

We run a reproducible diagnostic audit with two open-weight instruction-tuned model judges, DeepSeek-V4-Pro and Qwen/Qwen3.6-35B-A3B, accessed through third-party inference endpoints; open-weight status identifies the model, not the serving mode. Judges see blinded user-visible outputs for 90 held-out cases and three systems, plus a plain-English runtime-control reference. They do not see method names, automatic scores, raw production data, credentials, or oracle labels. Judges score six fidelity dimensions on a 0--2 rubric and choose the most faithful output per case. We use this as model-based fidelity evidence, not decision-quality evidence; model judges have known reliability and bias limits \cite{liu2023geval,zheng2023llmjudge,gu2024surveyjudge,ye2024judgebias}.

\begin{table}[H]
\centering
\scriptsize
\begingroup
\renewcommand{\arraystretch}{1.08}
\setlength{\tabcolsep}{2.8pt}
\begin{tabular}{lrrrr}
\toprule
\textbf{System} & \textbf{Overall} & \textbf{NFE} & \textbf{Surf.} & \textbf{Win} \\
\midrule
Raw & \textbf{1.6778} & 1.7667 & \textbf{1.8778} & 30.6\% \\
CBEA+LCV & 1.6509 & \textbf{1.7944} & 1.8167 & \textbf{50.0\%} \\
Validator & 1.3843 & 1.5278 & 1.7889 & 17.8\% \\
\bottomrule
\end{tabular}
\endgroup
\caption{Model-judge diagnostic. Overall, NFE (infeasible handling), and Surf.\ are dimension means on a 0--2 rubric; Win is the share of 180 case-winner selections across two judges and 90 cases, not human preference. \textbf{Bold} marks the column max. Detailed judge scores are in Table~\ref{tab:appendix-judge-details}; inter-judge agreement is in Table~\ref{tab:appendix-judge-agreement}.}
\label{tab:llm-judge-main}
\end{table}

The audit is conservative by design: dimension means remain close, with raw slightly higher on overall and surface coherence, while winner selections favor CBEA+LCV under both judges (Table~\ref{tab:llm-judge-main}). A case-cluster bootstrap gives CBEA+LCV a 50.0\% winner share with 95\% CI [41.7, 58.3] and a 19.4-point margin over raw with CI [5.0, 33.9] (Appendix~\ref{app:model-judge-details}). We treat this as secondary model-based fidelity evidence, not human preference.

\section{Discussion}

CBEA and LCV change the operating point of the evaluated runtime. CBEA makes evidence selection an explicit budgeted optimization; LCV makes commitment feasibility a lexicographic decision rather than a soft preference. Raw prompt stuffing remains a strong baseline because the benchmark exposes relevant evidence directly in the prompt, so the contribution is narrower and systems-oriented: runtime control failures become measurable, selected witnesses are protected under bounded activation, and infeasible states route to non-commitment acts.

The main comparison is therefore not ``short prompt beats long prompt.'' It is a trade between implicit and explicit control. Raw prompting gives the generator broad access to text but leaves the interpretation of that text to the model at each turn. CBEA+LCV narrows the active state, which loses some uncompiled context, but turns covered constraints, required evidence, witnesses, and repair states into objects that can be checked before prose. The shadow diagnostic is important precisely because it shows this cost rather than smoothing it away.

The long-history diagnostic gives the systems interpretation. With synthetic histories lengthened and runtime state held fixed, raw prompt stuffing carries the full archive while CBEA carries the selected evidence subset, reducing median input payload by 74--75\% (15.3k--16.3k tokens per case) across MiniMax, DeepSeek, and Qwen endpoints. Latency varies with provider behavior and output length, so input-payload reduction is the cross-endpoint result.

The evidence design separates deployment realism from comparative proof: production aggregates show scale and failure prevalence; fixtures anchor controlled stress surfaces without exposing user data; controlled experiments establish automatic runtime effects; and the model-judge audit is secondary diagnostic evidence, not human validation. The result is a systems and evaluation contribution, not a theory of language control or evidence of real-world decision improvement.

\section{Conclusion}

CBEA and LCV separate evidence activation from commitment validation in a long-horizon personalization runtime. Across matched, backend-sensitivity, long-history, shadow-boundary, and model-judge diagnostics, CBEA+LCV makes runtime control failures measurable and reaches zero measured failures within validator scope at substantially higher structured-commitment availability than validation applied to raw or long-context baselines. The contribution is a runtime-control operating point: a way to trade raw context breadth for explicit commitment control, with the loss of uncompiled context measured rather than hidden.

\section{Limitations}

The runtime guarantee is limited to explicitly confirmed structured commitments covered by validators. It does not certify unrestricted natural language semantics. Surface realization can still introduce ambiguity or unsupported implications after a valid commitment has been selected.

Production aggregates are observational: they establish deployment scale, reliability, and failure prevalence, not causal improvement or user-outcome quality. The benchmark is diagnostic, not distributional; results should not be read as average user-traffic performance. The fixtures intentionally isolate prespecified runtime-control stress surfaces rather than estimating their frequency in production traffic. Fixtures are synthetic or composite transformations of observed failure patterns, not raw user histories or a representative user sample.

The architecture depends on the quality of clarification, contract compilation, evidence labeling, and validator coverage; missing witnesses, false hardening, and incomplete validators remain possible failure modes. The fidelity audit is model-based and should be read as a reproducible diagnostic signal rather than human validation. We do not claim novelty in preference inference, memory consolidation, overpersonalization diagnosis, or contract-grounded verification in isolation; the contribution is their integration as a validator-covered runtime-control operating point.

\section{Ethical Considerations and Privacy}

Long-horizon personalized systems can create an illusion of certainty, especially in decision support settings. The runtime is designed to reduce one class of failure---drift across explicitly confirmed boundaries---but it does not justify medical, legal, financial, mental health, relationship, career, or life outcome claims.

The evaluation uses only privacy protected production aggregates and synthetic, composite, or heavily redacted fixtures. We exclude raw histories, raw biographies, raw core dilemmas, emails, user IDs, session tokens, exact payment records, exact timestamps tied to individuals, and identifiable life event combinations. The judge prompts and artifacts do not include credentials, method names for blinded outputs, automatic scores, raw production histories, or user identifiers.

\bibliography{nlp-systems-emnlp-arr-2026}

@misc{google_clarify_2025,
  title = {Asking Clarifying Questions for Preference Elicitation with Large Language Models},
  author = {Montazeralghaem, Ali and Tennenholtz, Guy and Boutilier, Craig and Meshi, Ofer},
  year = {2025},
  eprint = {2510.12015},
  archivePrefix = {arXiv},
  primaryClass = {cs.AI},
  url = {https://research.google/pubs/asking-clarifying-questions-for-preference-elicitation-with-large-language-models/}
}

@misc{selm2024,
  title = {Self-Exploring Language Models: Active Preference Elicitation for Online Alignment},
  author = {Zhang, Shenao and Yu, Donghan and Sharma, Hiteshi and Zhong, Han and Liu, Zhihan and Yang, Ziyi and Wang, Shuohang and Hassan, Hany and Wang, Zhaoran},
  year = {2024},
  eprint = {2405.19332},
  archivePrefix = {arXiv},
  primaryClass = {cs.LG},
  url = {https://arxiv.org/abs/2405.19332}
}

@inproceedings{personax2025,
  title = {{P}ersona{X}: A Recommendation Agent-Oriented User Modeling Framework for Long Behavior Sequence},
  author = {Shi, Yunxiao and Xu, Wujiang and Zeqi, Zhang and Zi, Xing and Wu, Qiang and Xu, Min},
  booktitle = {Findings of the Association for Computational Linguistics: ACL 2025},
  year = {2025},
  address = {Vienna, Austria},
  publisher = {Association for Computational Linguistics},
  pages = {5764--5787},
  doi = {10.18653/v1/2025.findings-acl.300},
  url = {https://aclanthology.org/2025.findings-acl.300/}
}

@inproceedings{difference_aware_2025,
  title = {Measuring What Makes You Unique: Difference-Aware User Modeling for Enhancing {LLM} Personalization},
  author = {Qiu, Yilun and Zhao, Xiaoyan and Zhang, Yang and Bai, Yimeng and Wang, Wenjie and Cheng, Hong and Feng, Fuli and Chua, Tat-Seng},
  booktitle = {Findings of the Association for Computational Linguistics: ACL 2025},
  year = {2025},
  address = {Vienna, Austria},
  publisher = {Association for Computational Linguistics},
  pages = {21258--21277},
  doi = {10.18653/v1/2025.findings-acl.1095},
  url = {https://aclanthology.org/2025.findings-acl.1095/}
}

@inproceedings{context_length_2025,
  title = {Context Length Alone Hurts {LLM} Performance Despite Perfect Retrieval},
  author = {Du, Yufeng and Tian, Minyang and Ronanki, Srikanth and Rongali, Subendhu and Bodapati, Sravan Babu and Galstyan, Aram and Wells, Azton and Schwartz, Roy and Huerta, Eliu A and Peng, Hao},
  booktitle = {Findings of the Association for Computational Linguistics: EMNLP 2025},
  year = {2025},
  address = {Suzhou, China},
  publisher = {Association for Computational Linguistics},
  pages = {23281--23298},
  doi = {10.18653/v1/2025.findings-emnlp.1264},
  url = {https://aclanthology.org/2025.findings-emnlp.1264/}
}

@misc{horizonbench2026,
  title = {HorizonBench: Long-Horizon Personalization with Evolving Preferences},
  author = {Li, Shuyue Stella and Paranjape, Bhargavi and Oktar, Kerem and Ma, Zhongyao and Zhou, Gelin and Guan, Lin and Zhang, Na and Park, Sem and Chen, Lin and Yang, Diyi and Tsvetkov, Yulia and Celikyilmaz, Asli},
  year = {2026},
  eprint = {2604.17283},
  archivePrefix = {arXiv},
  primaryClass = {cs.CL},
  url = {https://arxiv.org/abs/2604.17283}
}

@misc{cupid2025,
  title = {{CUPID}: Evaluating Personalized and Contextualized Alignment of {LLM}s from Interactions},
  author = {Kim, Tae Soo and Lee, Yoonjoo and Park, Yoonah and Kim, Jiho and Kim, Young-Ho and Kim, Juho},
  year = {2025},
  eprint = {2508.01674},
  archivePrefix = {arXiv},
  primaryClass = {cs.CL},
  url = {https://arxiv.org/abs/2508.01674},
  note = {Accepted to COLM 2025}
}

@misc{timem2026,
  title = {{TiMem}: Temporal-Hierarchical Memory Consolidation for Long-Horizon Conversational Agents},
  author = {Li, Kai and Yu, Xuanqing and Ni, Ziyi and Zeng, Yi and Xu, Yao and Zhang, Zheqing and Li, Xin and Sang, Jitao and Duan, Xiaogang and Wang, Xuelei and Liu, Chengbao and Tan, Jie},
  year = {2026},
  eprint = {2601.02845},
  archivePrefix = {arXiv},
  primaryClass = {cs.CL},
  url = {https://arxiv.org/abs/2601.02845}
}

@misc{opbench2026,
  title = {{OP-Bench}: Benchmarking Over-Personalization for Memory-Augmented Personalized Conversational Agents},
  author = {Hu, Yulin and Long, Zimo and Guo, Jiahe and Sui, Xingyu and Fu, Xing and Zhao, Weixiang and Zhao, Yanyan and Qin, Bing},
  year = {2026},
  eprint = {2601.13722},
  archivePrefix = {arXiv},
  primaryClass = {cs.CL},
  url = {https://arxiv.org/abs/2601.13722}
}

@inproceedings{carbonell1998mmr,
  title = {The Use of {MMR}, Diversity-Based Reranking for Reordering Documents and Producing Summaries},
  author = {Carbonell, Jaime and Goldstein, Jade},
  booktitle = {Proceedings of the 21st Annual International ACM SIGIR Conference on Research and Development in Information Retrieval},
  year = {1998},
  pages = {335--336},
  doi = {10.1145/290941.291025},
  url = {https://doi.org/10.1145/290941.291025}
}

@misc{contract2plan2026,
  title = {Contract2Plan: Verified Contract-Grounded Retrieval-Augmented Optimization for {BOM}-Aware Procurement and Multi-Echelon Inventory Planning},
  author = {Agarwal, Sahil},
  year = {2026},
  eprint = {2601.06164},
  archivePrefix = {arXiv},
  primaryClass = {cs.SE},
  url = {https://arxiv.org/abs/2601.06164}
}

@misc{liu2023geval,
  title = {{G-Eval}: {NLG} Evaluation using {GPT-4} with Better Human Alignment},
  author = {Liu, Yang and Iter, Dan and Xu, Yichong and Wang, Shuohang and Xu, Ruochen and Zhu, Chenguang},
  year = {2023},
  eprint = {2303.16634},
  archivePrefix = {arXiv},
  primaryClass = {cs.CL},
  url = {https://arxiv.org/abs/2303.16634}
}

@misc{zheng2023llmjudge,
  title = {Judging {LLM}-as-a-Judge with {MT-Bench} and Chatbot Arena},
  author = {Zheng, Lianmin and Chiang, Wei-Lin and Sheng, Ying and Zhuang, Siyuan and Wu, Zhanghao and Zhuang, Yonghao and Lin, Zi and Li, Zhuohan and Li, Dacheng and Xing, Eric P. and Zhang, Hao and Gonzalez, Joseph E. and Stoica, Ion},
  year = {2023},
  eprint = {2306.05685},
  archivePrefix = {arXiv},
  primaryClass = {cs.CL},
  url = {https://arxiv.org/abs/2306.05685}
}

@misc{gu2024surveyjudge,
  title = {A Survey on {LLM}-as-a-Judge},
  author = {Gu, Jiawei and Jiang, Xuhui and Shi, Zhichao and Tan, Hexiang and Zhai, Xuehao and Xu, Chengjin and Li, Wei and Shen, Yinghan and Ma, Shengjie and Liu, Honghao and Wang, Saizhuo and Zhang, Kun and Wang, Yuanzhuo and Gao, Wen and Ni, Lionel and Guo, Jian},
  year = {2024},
  eprint = {2411.15594},
  archivePrefix = {arXiv},
  primaryClass = {cs.CL},
  url = {https://arxiv.org/abs/2411.15594}
}

@misc{ye2024judgebias,
  title = {Justice or Prejudice? Quantifying Biases in {LLM}-as-a-Judge},
  author = {Ye, Jiayi and Wang, Yanbo and Huang, Yue and Chen, Dongping and Zhang, Qihui and Moniz, Nuno and Gao, Tian and Geyer, Werner and Huang, Chao and Chen, Pin-Yu and Chawla, Nitesh V. and Zhang, Xiangliang},
  year = {2024},
  eprint = {2410.02736},
  archivePrefix = {arXiv},
  primaryClass = {cs.CL},
  url = {https://arxiv.org/abs/2410.02736}
}

\appendix

\section{Reproducibility Artifact}
\label{app:artifact}

Table~\ref{tab:artifact-card} lists the anonymous review artifact for
reproducing reported diagnostics; it excludes author/repository identifiers,
credentials, raw production or user data, and exact user-linked timestamps.

\begin{table}[H]
\centering
\scriptsize
\begingroup
\renewcommand{\arraystretch}{1.08}
\setlength{\tabcolsep}{3pt}
\begin{tabularx}{\columnwidth}{@{}p{0.30\columnwidth}X@{}}
\toprule
\textbf{Checklist item} & \textbf{Artifact contents} \\
\midrule
Fixtures & Generator, JSON schema, and 360 released fixtures \\
Results & Model-output score tables, ablations, backend sensitivity, and long-history payload diagnostics \\
Metrics & Summarizers, denominator checks, release-table checks, and public table cases \\
Uncertainty & Paired bootstrap and case-cluster bootstrap scripts \\
Baselines & Matched harness, LCV-gated baselines, and selector-level MMR diagnostic \\
Judge audit & Blinded 90-case items, annotation key, labels, and winner selections \\
Privacy & Aggregate-only production diagnostics and privacy-boundary checks \\
\bottomrule
\end{tabularx}
\endgroup
\caption{Anonymous reproducibility artifact card.}
\label{tab:artifact-card}
\end{table}

\section{CBEA and LCV Reporting Details}
\label{app:method-details}

Table~\ref{tab:baselines} summarizes the comparison interfaces in the matched
offline run. Table~\ref{tab:method-details} reports the fixed method choices
used by CBEA+LCV. The values are design-prior settings, not learned parameters
or post-hoc tuning.

\begin{table}[H]
\centering
\scriptsize
\begingroup
\renewcommand{\arraystretch}{1.06}
\setlength{\tabcolsep}{2pt}
\begin{tabularx}{\columnwidth}{@{}p{0.30\columnwidth}p{0.15\columnwidth}X@{}}
\toprule
\textbf{Method} & \textbf{Ctrl.} & \textbf{Interface and purpose} \\
\midrule
Raw prompt stuffing & No/No & Full profile or memory text in prompt; naive personalization under budget pressure \\
\addlinespace[0.25em]
Summarized profile & No/No & LLM-compressed user history; product-style compression baseline \\
\addlinespace[0.25em]
Dense retrieval RAG & No/No & Embedding top-$k$ snippets; tests whether similarity retrieval is enough \\
\addlinespace[0.25em]
Long-context LLM & No/No & Maximum feasible history; tests whether longer context alone solves control \\
\addlinespace[0.25em]
Tool/memory agent & No/No & Write memory, retrieve memory, call tools; strong agent-memory baseline \\
\addlinespace[0.25em]
Validator-only & Yes/Yes & Structured candidates filtered by $V_h$; no CBEA activation \\
\addlinespace[0.25em]
Runtime without CBEA & Yes/Yes & Structured state and validator with non-tail evidence selection \\
\addlinespace[0.25em]
CBEA + LCV runtime & Yes/Yes & $h_t,E_t,u_t,R_t,M_t,Z_t,a_t$ with $\Repair$; main method \\
\addlinespace[0.25em]
Oracle evidence upper bound & Yes/Yes & Oracle-selected evidence witnesses; non-deployable upper bound \\
\bottomrule
\end{tabularx}
\endgroup
\caption{Comparison set for the matched offline run. Ctrl. is hard-contract/repair-path availability. Oracle evidence is a reference upper bound rather than a deployable baseline.}
\label{tab:baselines}
\end{table}

\begin{table}[H]
\centering
\scriptsize
\begingroup
\renewcommand{\arraystretch}{1.06}
\setlength{\tabcolsep}{2pt}
\begin{tabularx}{\linewidth}{@{}p{0.28\linewidth}X@{}}
\toprule
\textbf{Component} & \textbf{Reported implementation} \\
\midrule
CBEA weights & $(\lambda_r,\lambda_c,\lambda_w,\lambda_d,\lambda_o)=(1,2,2,1,1)$ for relevance, coverage, tail witness, consequence debt, and overuse penalty \\
\addlinespace[0.2em]
Budget policy & Greedy budgeted coverage with tail-witness reservation \\
\addlinespace[0.2em]
Requirement sources & Confirmed hard predicates; required dimensions and slots; runtime-detected missing requirements; tail-witness requirements; consequence obligations compiled from confirmed state; no-feasible guards derived by runtime rules \\
\addlinespace[0.2em]
Coverage matrix & Binary evidence-unit by requirement matrix; $M_{ir}=1$ when evidence unit $e_i$ satisfies requirement $r$ \\
\addlinespace[0.2em]
Candidate schema & Selected option; commitment type; hard predicates used; evidence witness ids; covered requirements; consequence obligations; repair or abstain reason; surface requirements \\
\addlinespace[0.2em]
Validator predicates & Hard-contract consistency; required-evidence coverage; no-feasible emission guard; repair reason validity; surface-requirement presence \\
\addlinespace[0.2em]
LCV ordering & Hard predicate violations, then coverage violations, then no-feasible emission, then soft score \\
\bottomrule
\end{tabularx}
\endgroup
\caption{Fixed CBEA/LCV reporting details for the matched offline run.}
\label{tab:method-details}
\end{table}

\FloatBarrier
\section{Production Aggregate Details}
\label{app:production}

Table~\ref{tab:production} gives the production data-wash denominators used for
deployment-scale diagnostics.

\begin{table}[H]
\centering
\scriptsize
\begingroup
\renewcommand{\arraystretch}{1.04}
\setlength{\tabcolsep}{2pt}
\begin{tabularx}{\linewidth}{@{}p{0.31\linewidth}rp{0.18\linewidth}X@{}}
\toprule
\textbf{Surface} & \textbf{Count} & \textbf{Rate} & \textbf{Use} \\
\midrule
Attempted sessions & 524 & 100.0\% & Deployment denominator \\
Semantic session pool & 447 & 85.3\% & After data wash \\
Completed sessions & 398 & 76.0\% & Funnel state \\
Final evaluable sessions & 323 & 61.6\% & Result analysis \\
Active/unfinished & 98 & 18.7\% & Not correct outputs \\
Operational invalid & 77 & 14.7\% & Excluded from semantics \\
Evaluable turns & 6,018 & 99.0\% & Turn denominator \\
Provider requests & 28,815 & 100.0\% & Request denominator \\
Provider successes & 19,988 & 69.4\% & Reliability \\
\bottomrule
\end{tabularx}
\endgroup
\caption{Privacy-safe production data-wash export. These values support deployment diagnostics only.}
\label{tab:production}
\end{table}

Runtime-control fields were populated for most semantic sessions: runtime
context for 445 of 447, contract-state markers for 396, required dimensions for
391, selected dimensions for 396, and consequence debt for 323.

\paragraph{Denominator discipline.}
\label{app:denominators}

Evaluation scripts use the same denominator discipline as reporting. Production
sessions define deployment scale, task mix, completion, error rate, provider
reliability, latency, and runtime activation; completed/evaluable outputs are
used only for readiness summaries. Attempted benchmark runs define pass,
timeout, no-output, and no-feasible handling rates; benchmark outputs with text
define fidelity labels and deterministic proxy validation. Unfinished, errored,
partial, blank, provider-side, and parsing-failure states are reliability
outcomes, not correct semantic outputs.

\paragraph{Fixture and data boundaries.}
\label{app:fixture-boundaries}

The controlled fixtures are synthetic or composite and preserve seven stress
surfaces. The 360-fixture manifest exports six primary buckets with 60 fixtures
each (Table~\ref{tab:fixture-buckets}); over-personalization is evaluated as an
activation penalty and diagnostic rather than a separate bucket. Fixtures exclude
raw production histories, biographies, session identifiers, payment records,
exact user-linked timestamps, and identifiable life-event combinations.

\begin{table}[H]
\centering
\footnotesize
\begingroup
\renewcommand{\arraystretch}{1.08}
\setlength{\tabcolsep}{3pt}
\begin{tabular}{@{}lrl@{}}
\toprule
\textbf{Bucket} & \textbf{N} & \textbf{Stress surface} \\
\midrule
falsehard & 60 & False hardening \\
exception & 60 & Hidden exception \\
tail & 60 & Witness drop \\
infeasible & 60 & No feasible candidate \\
debt & 60 & Consequence debt \\
surface & 60 & Surface mismatch \\
\bottomrule
\end{tabular}
\endgroup
\caption{Primary exported fixture buckets in the matched 360-fixture run. Counts are over unique fixture identifiers in the merged evaluation manifest; each fixture is evaluated under all nine variants.}
\label{tab:fixture-buckets}
\end{table}

\begin{table}[H]
\centering
\scriptsize
\begingroup
\renewcommand{\arraystretch}{1.06}
\setlength{\tabcolsep}{3pt}
\begin{tabularx}{\columnwidth}{@{}p{0.30\columnwidth}X@{}}
\toprule
\textbf{Stress surface} & \textbf{Primary metric and test} \\
\midrule
False hardening & False-hardening rate; whether noisy hints become hard constraints \\
\addlinespace[0.25em]
Hidden exception & OHCVR; whether validation respects exception scope \\
\addlinespace[0.25em]
Witness drop & Evidence coverage failure; whether activation preserves rare evidence \\
\addlinespace[0.25em]
No feasible candidate & NFER and repair correctness; whether the system emits or repairs/abstains \\
\addlinespace[0.25em]
Consequence debt & Consequence continuity failure; whether state carries downstream obligations \\
\addlinespace[0.25em]
Over-personalization & Personalization diagnostic; whether irrelevant memory is suppressed \\
\addlinespace[0.25em]
Surface mismatch & Surface coherence audit; whether structure survives prose realization \\
\bottomrule
\end{tabularx}
\endgroup
\caption{Benchmark stress surfaces. Each fixture has structured labels, so failure metrics are defined over runtime objects rather than generic one-turn quality.}
\label{tab:benchmark}
\end{table}

\section{Long-History Prompt-Payload Diagnostic}
\label{app:long-history-payload}

Table~\ref{tab:appendix-long-history-payload} reports a separate 50-fixture
payload diagnostic. Only the user-visible history archive is lengthened:
confirmed contracts, compiled requirements, and runtime rules are unchanged.
Raw receives the full history; CBEA+LCV receives the evidence subset selected by
the harness. The runner records provider usage and wall-clock time, but does not
parse outputs or perform JSON retries.

\begin{table}[H]
\centering
\tiny
\begingroup
\renewcommand{\arraystretch}{1.03}
\setlength{\tabcolsep}{2pt}
\resizebox{\columnwidth}{!}{%
\begin{tabular}{lrrrrrrr}
\toprule
\textbf{Endpoint} & \textbf{Raw in p50} & \textbf{CBEA in p50} & \textbf{$\Delta$ in p50} & \textbf{Raw out p50} & \textbf{CBEA out p50} & \textbf{$\Delta$ lat mean} & \textbf{$\Delta$ lat p50} \\
\midrule
MiniMax & 20,401 & 5,154 & 15,258 & 700 & 700 & +1.74s & +1.54s \\
DeepSeek & 21,364 & 5,484 & 15,929 & 700 & 700 & +2.06s & +1.97s \\
Qwen & 21,709 & 5,376 & 16,279 & 2,194 & 2,657 & -0.80s & -1.33s \\
\bottomrule
\end{tabular}}
\endgroup
\caption{Long-history prompt-payload diagnostic. Inputs and outputs are provider-reported tokens. $\Delta$ is raw prompt stuffing minus CBEA+LCV; positive latency means CBEA was faster within that endpoint. The supported cross-endpoint result is input payload reduction, not latency.}
\label{tab:appendix-long-history-payload}
\end{table}

Median input falls by 74--75\% (15.3k--16.3k tokens per case). Mean latency
intervals are favorable on MiniMax (+1.74s, 95\% CI [1.18, 2.29]) and DeepSeek
(+2.06s, [1.45, 2.69]); Qwen keeps the same input reduction but has longer CBEA
completions and a mean latency interval crossing zero (-0.80s, [-2.01, 0.42]).
We therefore treat input payload reduction as the supported cross-endpoint
result and latency as endpoint/output-length dependent.

\begin{table}[H]
\centering
\scriptsize
\begingroup
\renewcommand{\arraystretch}{1.06}
\setlength{\tabcolsep}{4pt}
\resizebox{\columnwidth}{!}{%
\begin{tabular}{lrr}
\toprule
\textbf{Surface} & \textbf{Diff.} & \textbf{95\% CI} \\
\midrule
Structured availability & $-0.2833$ & $[-0.3361,-0.2306]$ \\
Parse retries per fixture & $+0.0333$ & $[-0.0250,0.0944]$ \\
All-row repair correctness & \textbf{$+0.5429$} & \textbf{[0.4571,0.6286]} \\
Observed short-prompt latency, ms & $+1141$ & $[-882,3132]$ \\
\bottomrule
\end{tabular}
}
\endgroup
\caption{Paired bootstrap intervals over 360 fixtures for CBEA+LCV minus raw prompt stuffing. Positive is better for availability and repair correctness; negative is better for retries and short-prompt latency. \textbf{Bold} marks intervals excluding zero. The latency row is separate from the long-history payload diagnostic in Table~\ref{tab:appendix-long-history-payload}. Negative structured availability reflects LCV routing infeasible fixtures to non-commitment control acts rather than unsupported commitments.}
\label{tab:appendix-bootstrap}
\end{table}

\section{Horizon-Complexity Stability Diagnostic}
\label{app:horizon-stability}

The fixtures do not contain real turn-depth labels, so we do not report a
turn-depth causal analysis. As a conservative proxy, Table~\ref{tab:appendix-horizon-stability}
groups the matched run by the number of required domains in each fixture. The
split reuses the same 360-fixture outputs and adds no new model calls. CBEA+LCV
keeps oracle-level OHCVR, witness-drop, and consequence-continuity failures at
zero in every group, while repair correctness remains 1.0. Parse retries are
mixed and are not used as a main claim. Oracle-level OHCVR remains an end-to-end
fidelity diagnostic rather than a validator-covered invariant.

\begin{table}[H]
\centering
\tiny
\begingroup
\renewcommand{\arraystretch}{1.03}
\setlength{\tabcolsep}{2pt}
\resizebox{\columnwidth}{!}{%
\begin{tabular}{llrrrrrrr}
\toprule
\textbf{Group} & \textbf{Method} & \textbf{Fix.} & \textbf{Struct.} & \textbf{OHCVR} & \textbf{Wit.} & \textbf{Cons.} & \textbf{Repair} & \textbf{Retry} \\
\midrule
2-domain & Raw & 144 & 119 & 0.1176 & 0.0672 & 0.4622 & 0.4286 & 0.13 \\
2-domain & CBEA+LCV & 144 & 80 & 0.0000 & 0.0000 & 0.0000 & 1.0000 & 0.15 \\
3-domain & Raw & 144 & 106 & 0.1792 & 0.0943 & 0.4623 & 0.5000 & 0.15 \\
3-domain & CBEA+LCV & 144 & 61 & 0.0000 & 0.0000 & 0.0000 & 1.0000 & 0.24 \\
4-domain & Raw & 72 & 57 & 0.0526 & 0.0175 & 0.3860 & 0.4286 & 0.19 \\
4-domain & CBEA+LCV & 72 & 39 & 0.0000 & 0.0000 & 0.0000 & 1.0000 & 0.14 \\
\bottomrule
\end{tabular}}
\endgroup
\caption{Horizon-complexity diagnostic over the matched 360-fixture run. Groups are based on the number of required domains in each fixture, not real conversation turn depth. OHCVR, Wit., and Cons. are lower-is-better diagnostic rates; Repair is higher-is-better repair or abstention correctness; Retry is mean parse retries per fixture.}
\label{tab:appendix-horizon-stability}
\end{table}

\section{LCV Repair Case Sketch}
\label{app:lcv-repair-case}

Table~\ref{tab:appendix-lcv-repair-case} gives a qualitative sketch from one
relocation-career fixture. The confirmed contract includes
no new unsecured debt, preserving a weekly caregiving obligation, and not
asserting partner approval without a witness. Activated evidence contains the
offer constraints, debt boundary, and caregiving schedule; partner backup is
missing. The example is not a decision quality case study. It shows how LCV
routes infeasible or unsupported states before prose realization.

\begin{table}[H]
\centering
\scriptsize
\begingroup
\renewcommand{\arraystretch}{1.08}
\setlength{\tabcolsep}{2pt}
\begin{tabularx}{\columnwidth}{@{}>{\raggedright\arraybackslash}p{0.24\columnwidth}YY@{}}
\toprule
\textbf{LCV diagnosis} & \textbf{Detected condition} & \textbf{Repair action} \\
\midrule
Missing evidence & Candidate assumes backup care or partner approval without an activated witness. & Clarify the missing premise; no recommendation is emitted. \\
\addlinespace[0.2em]
Contract conflict & Candidate requires protected caregiving time or new debt, contradicting $h_t$. & Recontract: state the conflict and ask whether to revise the boundary; do not emit the infeasible commitment. \\
\addlinespace[0.2em]
Unsupported commitment & Candidate adds household approval or debt-safety claims not supported by evidence. & Abstain or fallback to a verified comparison with explicit uncertainty. \\
\bottomrule
\end{tabularx}
\endgroup
\caption{Illustrative LCV repair routing for one synthetic and composite fixture. The table distinguishes clarification, recontract, and abstention or fallback as separate non-commitment control acts.}
\label{tab:appendix-lcv-repair-case}
\end{table}

\section{Targeted Runtime Ablations}
\label{app:targeted-ablations}

Table~\ref{tab:appendix-targeted-ablations} reports targeted ablations from a
separate 360-fixture component-diagnostic run. The comparison is diagnostic: it tests
whether removing a runtime component degrades the surface that component is
designed to control, rather than replacing the main matched comparison in Table~\ref{tab:offline-results}.
Absolute values should be read as targeted component diagnostics rather than as
additional baseline rows for Table~\ref{tab:offline-results}.

\begin{table}[H]
\centering
\tiny
\begingroup
\renewcommand{\arraystretch}{1.03}
\setlength{\tabcolsep}{2pt}
\resizebox{\columnwidth}{!}{%
\begin{tabular}{lrrrrrrrr}
\toprule
\textbf{Variant} & \textbf{Inv.} & \textbf{Struct.} & \textbf{OHCVR} & \textbf{ECF} & \textbf{Wit.} & \textbf{Cons.} & \textbf{NFER} & \textbf{Rep.} \\
\midrule
CBEA+LCV diagnostic & 0 & 0.8361 & \textbf{0.0631} & \textbf{0.0532} & \textbf{0.0465} & 0.0498 & \textbf{0.0066} & \textbf{0.9508} \\
No validator & 0 & 0.8583 & 0.1521 & 0.1489 & 0.1553 & 0.0971 & 0.0680 & 0.5417 \\
No repair/abstain & 0 & 1.0000 & 0.2000 & 0.0222 & 0.0250 & 0.0139 & 0.1667 & 0.0000 \\
No coverage/tail & 0 & 0.8167 & 0.2143 & 1.0000 & 1.0000 & 1.0000 & 0.0068 & 0.8657 \\
\bottomrule
\end{tabular}}
\endgroup
\caption{Targeted ablation results from a separate 360-fixture component diagnostic. Lower is better for Inv., OHCVR, ECF, Wit., Cons., and NFER; higher is better for Struct. and Rep. \textbf{Bold} marks intended control surfaces where the full diagnostic runtime improves over targeted removals. This run is not a replacement for the matched comparison in Table~\ref{tab:offline-results}.}
\label{tab:appendix-targeted-ablations}
\end{table}

\section{Validator-Covered Control Failures: 9-Method Detail}
\label{app:covered-detail}

Table~\ref{tab:appendix-covered-detail} expands the matched nine-method comparison in Table~\ref{tab:offline-results}. LCV suppresses infeasible commitments by construction, but validation alone does not guarantee evidence coverage, witness retention, or consequence continuity; CBEA+LCV is the only non-oracle row with zeros across all covered metrics.

\begin{table}[H]
\centering
\scriptsize
\begingroup
\renewcommand{\arraystretch}{1.05}
\setlength{\tabcolsep}{2.5pt}
\resizebox{\columnwidth}{!}{%
\begin{tabular}{lrrrrrr}
\toprule
\textbf{Method} & \textbf{Struct.} & \textbf{OHCVR} & \textbf{ECF} & \textbf{Wit.} & \textbf{Cons.} & \textbf{NFER} \\
\midrule
Raw prompt stuffing  & 0.7833 & 0.1277 & 0.4113 & 0.0674 & 0.4468 & 0.0333 \\
Summarized profile   & 0.4500 & 0.2469 & 1.0000 & 0.5988 & 0.9938 & 0.0333 \\
Dense retrieval RAG  & 0.7333 & 0.1364 & 0.2803 & 1.0000 & 1.0000 & 0.3833 \\
Long-context LLM     & 0.7278 & 0.1565 & 0.5344 & 0.0649 & 0.5458 & 0.0333 \\
Tool/memory agent    & 0.7500 & 0.1704 & 0.4407 & 0.6926 & 0.6074 & 0.0333 \\
\midrule
Validator-only       & 0.4028 & 0.2759 & 0.4621 & 1.0000 & 1.0000 & 0$^\dagger$ \\
Runtime w/o CBEA     & 0.5000 & 0.2444 & 0.4889 & 0.7778 & 0.5222 & 0$^\dagger$ \\
CBEA + LCV runtime   & 0.5000 & 0 & 0 & 0 & 0 & 0$^\dagger$ \\
\midrule
Oracle evidence (ref.) & 0.6111 & 0 & 0 & 0 & 0 & 0 \\
\bottomrule
\end{tabular}}
\endgroup
\caption{Validator-covered control failures on the matched 360-fixture MiniMax-M2.7 run. OHCVR: hard-constraint violation; ECF: evidence-coverage failure; Wit.: witness-drop rate; Cons.: consequence-continuity failure; NFER: no-feasible emission rate. $^\dagger$NFER $=0$ by Property~1. Validator-only and Runtime w/o CBEA show that LCV gating alone does not yield the CBEA+LCV zeros.}
\label{tab:appendix-covered-detail}
\end{table}

\section{Selector-Level MMR Diagnostic}
\label{app:mmr-selector}

Table~\ref{tab:appendix-mmr-selector} compares CBEA activation with a classic
relevance-diversity MMR selector on the same 360 fixtures and 12-unit evidence
budget. This selector-only diagnostic asks whether relevance-diversity alone
recovers the typed control evidence that LCV later needs. MMR uses local
scenario and observation tokens as the query and has no access to coverage,
tail-witness, or consequence-debt terms.

\begin{table}[H]
\centering
\scriptsize
\begingroup
\renewcommand{\arraystretch}{1.05}
\setlength{\tabcolsep}{2.5pt}
\resizebox{\columnwidth}{!}{%
\begin{tabular}{lrrrrrr}
\toprule
\textbf{Selector} & \textbf{Avg. $|Z|$} & \textbf{Hard} & \textbf{ReqW} & \textbf{Tail} & \textbf{Debt} & \textbf{Control} \\
\midrule
CBEA selector & 10.67 & \textbf{1.0000} & \textbf{1.0000} & \textbf{1.0000} & \textbf{0.9889} & \textbf{0.9970} \\
MMR relevance-diversity & 12.00 & \textbf{1.0000} & 0.9444 & 0.6667 & 0.0000 & 0.6960 \\
\bottomrule
\end{tabular}
}
\endgroup
\caption{Selector-level MMR diagnostic over the same fixture evidence universe and budget, without model generation. Columns report mean recall for hard constraints, required witnesses (ReqW), tail witnesses, consequence debt, and their union (Control).}
\label{tab:appendix-mmr-selector}
\end{table}

\section{Backend Sensitivity and Operating Points}
\label{app:backend-robustness}

Table~\ref{tab:appendix-backend-robustness} reruns the 360-fixture comparison on
MiniMax-M2.7, DeepSeek-V4-Flash, and GPT-OSS-120B, applying the same LCV gate to
Raw, Long-ctx, and CBEA+LCV. This is a validator-scope sensitivity diagnostic,
not robustness evidence or a second model-judge audit; Hy3-preview is separate
because it fails to emit parseable commitments at the matched 2200-token budget.

We report structured-output cost separately from covered correctness:
\textbf{Inv.} is invalid runs, \textbf{Bdg.} budget-exhausted JSON parse failure,
and \textbf{Long.} output at or above the 2200-token budget. All emitted
commitments have zero validator-covered failures, but availability differs:
CBEA+LCV reaches 0.4944, 0.6000, and 0.5056, while LCV-gated raw and long-context
baselines reach the same failure level only at 0.0028--0.0917 availability.

\begin{table}[H]
\centering
\scriptsize
\begingroup
\renewcommand{\arraystretch}{1.05}
\setlength{\tabcolsep}{2pt}
\resizebox{\columnwidth}{!}{%
\begin{tabular}{llrrrrrrrr}
\toprule
\textbf{Backend} & \textbf{Method} & \textbf{Inv.} & \textbf{Bdg.} & \textbf{Long.} & \textbf{Avail.} & \textbf{OHCVR} & \textbf{ECF} & \textbf{Wit.} & \textbf{Cons.} \\
\midrule
MiniMax-M2.7    & Raw + LCV       & 5 & 5 & 275 & 0.0361 & 0 & 0 & 0 & 0 \\
MiniMax-M2.7    & Long-ctx + LCV  & 1 & 1 & 287 & 0.0500 & 0 & 0 & 0 & 0 \\
MiniMax-M2.7    & CBEA + LCV      & 1 & 1 & 211 & \textbf{0.4944} & 0 & 0 & 0 & 0 \\
\midrule
DeepSeek-V4-Flash & Raw + LCV       & 0 & 0 & 0   & 0.0917 & 0 & 0 & 0 & 0 \\
DeepSeek-V4-Flash & Long-ctx + LCV  & 0 & 0 & 0   & 0.0778 & 0 & 0 & 0 & 0 \\
DeepSeek-V4-Flash & CBEA + LCV      & 0 & 0 & 0   & \textbf{0.6000} & 0 & 0 & 0 & 0 \\
\midrule
GPT-OSS-120B    & Raw + LCV       & 0 & 0 & 7   & 0.0028 & 0 & 0 & 0 & 0 \\
GPT-OSS-120B    & Long-ctx + LCV  & 0 & 0 & 9   & 0.0056 & 0 & 0 & 0 & 0 \\
GPT-OSS-120B    & CBEA + LCV      & 1 & 0 & 15  & \textbf{0.5056} & 0 & 0 & 0 & 0 \\
\bottomrule
\end{tabular}}
\endgroup
\caption{Backend sensitivity over 360 fixtures $\times$ 3 methods $\times$ 3 backends (3{,}240 attempted rows) at a 2200-token budget. \textbf{Inv.}: invalid runs; \textbf{Bdg.}: budget-exhausted JSON parse failures; \textbf{Long.}: outputs at or above budget; \textbf{Avail.}: attempted-run structured-commitment availability. Covered metrics are on emitted commitments and are zero in every cell; \textbf{bold} marks the highest availability at each backend.}
\label{tab:appendix-backend-robustness}
\end{table}

\section{Hy3-preview Output-Budget Diagnostic}
\label{app:hy3-diag}

Hy3-preview (Hunyuan3-preview) is not backend-comparable under the matched
2200-token budget: the formal run was stopped after 64/64 budget-exhausted JSON
parse failures, and a low-concurrency check reproduced 3/3 invalid attempts. At
8800 tokens, Hy3-preview emits parseable commitments: one CBEA+LCV sample parses
and validates, and a 12-fixture $\times$ 3-method diagnostic yields 27/27 valid
runs before stopping. We treat this as an output-budget diagnostic only. It shows
that reasoning-heavy endpoints can spend the matched budget before final JSON;
the 8800-token run is small and not matched to the 360-fixture protocol.

\section{Shadow-Oracle Boundary Diagnostic}
\label{app:shadow-oracle}

The shadow-oracle layer evaluates what falls outside validator scope. Each
fixture includes hidden visible-fact labels, aliases, paraphrase patterns,
weights, and turn-due markers; runtime prompts receive only user-visible
observations and compiled fixture fields. Grep checks confirm that shadow fields
never appear in runtime prompts. We score uncompiled-fact recall by deterministic
alias and paraphrase matching over normalized output. Because CBEA selects only
from compiled evidence $E_t$, this MiniMax-only diagnostic measures selector
scope, not backend sensitivity.

\begin{table}[H]
\centering
\scriptsize
\begingroup
\renewcommand{\arraystretch}{1.05}
\setlength{\tabcolsep}{2.5pt}
\resizebox{\columnwidth}{!}{%
\begin{tabular}{lrrrrrr}
\toprule
\textbf{Method} & \textbf{Avail.} & \textbf{Cov.H} & \textbf{Cov.E} & \textbf{Cov.W} & \textbf{Cov.C} & \textbf{Uncomp.} \\
\midrule
Raw                & 0.7889 & 0.1901 & 0.5246 & 0.1972 & 0.8415 & 0.5316 \\
Raw + LCV          & 0.0361 & 0      & 0      & 0      & 0      & 0.3674\textsuperscript{*} \\
Long-ctx + LCV     & 0.0500 & 0      & 0      & 0      & 0      & 0.2380\textsuperscript{*} \\
Validator-only     & 0.4028 & 0.2000 & 0.4276 & 1.0000 & 1.0000 & 0.0000 \\
CBEA + LCV         & \textbf{0.4944} & 0 & 0 & 0 & 0 & \textbf{0.0119} \\
\bottomrule
\end{tabular}}
\endgroup
\caption{Shadow-oracle boundary diagnostic over 360 MiniMax-M2.7 fixtures. Avail. uses attempted denominator; covered metrics are on emitted commitments. \textbf{Uncomp.}: uncompiled-fact recall. \textsuperscript{*}Sparse denominator. CBEA+LCV preserves the compiled boundary at near-zero uncompiled recall; raw retains more uncompiled context while exhibiting high covered-boundary failures.}
\label{tab:appendix-shadow-overall}
\end{table}

\begin{table}[H]
\centering
\scriptsize
\begingroup
\renewcommand{\arraystretch}{1.05}
\setlength{\tabcolsep}{2.5pt}
\begin{tabular}{lrrrrr}
\toprule
\textbf{Domain} & \textbf{Raw} & \textbf{Raw+LCV} & \textbf{Long+LCV} & \textbf{Val-only} & \textbf{CBEA+LCV} \\
\midrule
investment      & 0.5733 & --     & 0      & 0 & 0.0098 \\
love\_choice    & 0.5246 & 0.5193 & 0.2292 & 0 & 0.0053 \\
career          & 0.4655 & 0.5000 & 0.2456 & 0 & 0.0132 \\
relocation      & 0.5479 & 0      & 0      & 0 & 0.0133 \\
comprehensive   & 0.5520 & 0.2222 & 0.5000 & 0 & 0.0187 \\
\bottomrule
\end{tabular}
\endgroup
\caption{Per-domain uncompiled-fact recall (MiniMax-M2.7 shadow-oracle, 360 fixtures). CBEA+LCV is near zero across all five domains, confirming the selector blindness is structural and not domain-specific. Raw shows non-trivial cross-domain variance (0.47--0.57), confirming the matcher is not domain-saturated.}
\label{tab:appendix-shadow-domain}
\end{table}

\section{Model-Judge Audit Details}
\label{app:model-judge-details}

Table~\ref{tab:appendix-judge-details} reports the open-weight model-judge audit
in more detail. Judges received blinded system labels, the same rubric, and a
plain-English runtime-control reference, but did not receive method names,
automatic scores, raw production data, credentials, or oracle labels. This audit
queried DeepSeek-V4-Pro and Qwen/Qwen3.6-35B-A3B through third-party inference
endpoints. Both endpoints were used only to obtain blinded model-judge outputs.
The model identifiers were \texttt{deepseek-v4-pro} and
\texttt{qwen3.6-35b-a3b}. We report the identifiers, decoding settings, and
judge prompts for auditability; exact outputs may vary with provider serving
implementations. Judge calls used temperature 0, top-$p$ 1, JSON response
format, the fixed 0--2 rubric, and the same per-case winner-selection prompt
for every held-out case.

Table~\ref{tab:appendix-judge-bootstrap} reports a case-cluster bootstrap over
the 90 held-out cases, resampling cases with replacement and preserving both
judge selections for each sampled case.
Per-dimension means are descriptive: raw prompt stuffing receives higher
surface-oriented scores on several dimensions, so Table~\ref{tab:appendix-judge-details}
is not interpreted as a method-winner table.

\begin{table}[H]
\centering
\scriptsize
\begingroup
\renewcommand{\arraystretch}{1.05}
\setlength{\tabcolsep}{3pt}
\begin{tabular}{lrrr}
\toprule
\textbf{Statistic} & \textbf{Observed} & \textbf{95\% CI low} & \textbf{95\% CI high} \\
\midrule
CBEA+LCV winner share & 0.5000 & 0.4167 & 0.5833 \\
Raw winner share & 0.3056 & 0.2278 & 0.3833 \\
Validator winner share & 0.1778 & 0.1111 & 0.2500 \\
Tie share & 0.0167 & 0.0000 & 0.0389 \\
CBEA--Raw margin & 0.1944 & 0.0500 & 0.3389 \\
\bottomrule
\end{tabular}
\endgroup
\caption{Case-cluster bootstrap over 180 model-judge winner selections from 90 cases and two judges. The CBEA--Raw margin is the difference in winner-selection share.}
\label{tab:appendix-judge-bootstrap}
\end{table}

\begin{table}[H]
\centering
\footnotesize
\begingroup
\renewcommand{\arraystretch}{1.04}
\setlength{\tabcolsep}{3.5pt}
\begin{tabular}{@{}llrrrr@{}}
\toprule
\textbf{Judge} & \textbf{System} & \textbf{Overall} & \textbf{Evid.} & \textbf{Cons.} & \textbf{Surf.} \\
\midrule
\textsc{DS} & CBEA & 1.6482 & 1.1333 & 1.3667 & 1.8333 \\
\textsc{DS} & Raw & 1.6704 & 1.2667 & 1.3889 & 1.8778 \\
\textsc{DS} & Val. & 1.4444 & 0.9111 & 0.8111 & 1.8333 \\
\addlinespace[0.2em]
\textsc{QW} & CBEA & 1.6537 & 1.3111 & 1.3222 & 1.8000 \\
\textsc{QW} & Raw & 1.6852 & 1.3444 & 1.3444 & 1.8778 \\
\textsc{QW} & Val. & 1.3241 & 0.7222 & 0.5111 & 1.7444 \\
\bottomrule
\end{tabular}
\endgroup
\caption{Per-judge descriptive means on a 0--2 rubric. \textsc{DS} is DeepSeek-V4-Pro, \textsc{QW} is Qwen/Qwen3.6-35B-A3B, CBEA is CBEA+LCV, Raw is raw prompt stuffing, and Val. is validator-only. These means are not the primary fidelity statistic and are not interpreted as method winners; they expose that raw prompt stuffing often receives higher surface/detail scores, while Table~\ref{tab:appendix-judge-bootstrap} reports the paired case-level winner diagnostic. Scores are not compared across judges.}
\label{tab:appendix-judge-details}
\end{table}

\begin{table}[H]
\centering
\footnotesize
\begingroup
\renewcommand{\arraystretch}{1.04}
\setlength{\tabcolsep}{4pt}
\resizebox{\columnwidth}{!}{%
\begin{tabular}{lrrr}
\toprule
\textbf{Dimension} & \textbf{N} & \textbf{Exact} & \textbf{Within 1} \\
\midrule
Constraint fidelity & 270 & 0.9222 & 0.9741 \\
Evidence coverage & 270 & 0.6593 & 0.9556 \\
Consequence continuity & 270 & 0.6630 & 0.9630 \\
No-feasible handling & 270 & 0.8111 & 0.9444 \\
Appropriate personalization & 270 & 0.9296 & 0.9741 \\
Surface coherence & 270 & 0.9185 & 0.9889 \\
All dimension labels & 1620 & 0.8173 & 0.9667 \\
Case-level winner & 90 & 0.6000 & -- \\
\bottomrule
\end{tabular}
}
\endgroup
\caption{Agreement between the two model judges. Exact agreement is strict equality on 0--2 labels or case-level winner.}
\label{tab:appendix-judge-agreement}
\end{table}

\end{document}